\documentclass[a4paper,10pt,twocolumn]{article}
\usepackage[utf8x]{inputenc}
\usepackage{mathptmx}
\usepackage{helvet}
\usepackage{courier}
\usepackage{amsmath,amsfonts}
\usepackage{graphicx}
\usepackage[bottom]{footmisc}
\usepackage[numbers]{natbib}    

\usepackage{a4wide}

\usepackage{algorithmic}
\usepackage{algorithm}

\usepackage{colortbl}

\usepackage{graphicx}

\title{Cable-Driven Robots with Wireless Control Capability for Pedagogical Illustration in Science}
\author{Julien Alexandre dit Sandretto$^1$, Cyprien Nicolas$^2$, \\
  Inria Sophia-Antipolis M\'editerran\'ee}
\date{May 27, 2013}

\begin{document}

\twocolumn[
  \begin{@twocolumnfalse}
    \maketitle
    \begin{abstract}

Science teaching in secondary schools is often abstract for
students. Even if some experiments can be conducted in classrooms,
mainly for chemistry or some physics fields, mathematics is not an
experimental science. Teachers have to convince students that theorems
have practical implications.
We present teachers an original and easy-to-use pedagogical tool:
a cable-driven robot with a Web-based remote control interface. The
robot implements several scientific concepts such as 3D-geometry and
kinematics. The remote control enables the teacher to move freely in
the classroom.

    \end{abstract}
    \bigskip 
  \end{@twocolumnfalse}
  ]

\footnotetext[1]{Coprin, julien.alexandre\_dit\_sandretto@inria.fr}
\footnotetext[2]{Indes, cyprien.nicolas@inria.fr}

\section{Introduction}
Students often find the instruction of technological sciences and
mathematics abstract and unclear. To illustrate their lecture and
explain theorems, teachers are restricted to use a limited number of
tools (\textit{e.g.} ruler, board) that have an abstract link with the concepts
that have to be illustrated. Thus the understanding is difficult, and
the interest of mathematics is not physically demonstrated.
Our recent work \cite{merlet_marionet_2010} led us to consider that
parallel cable-driven robot may be interesting as a pedagogical
demonstrator. Cable-driven robots use motorized winches to coil and
uncoil cables and a coordinated control of the cable lengths allows
one to control the position of a platform moving in space, like
a crane with several cables \cite{tadokoro_a_1999}.  Moreover,
teachers need to move around the robot and walk in the classroom to
physically illustrate their speech and provide an interesting
course. Under this constraint, the controller has to be remote
and wireless.

Cable-driven robots may be used to illustrate a large variety of
scientific concepts, such as geometry: Pythagorean theorem, vectors,
Grassmann geometry,~\textit{etc}.; robotics: control, calibration,
kinematics, statics,~\textit{etc}.; and computer science: user
interfaces, network communication, numerical methods,
programming,~\textit{etc}., this list being non exhaustive.

We present a pedagogical system that has already been used at a local
science fair, and we base our talk on the experience we gained from
that demonstration.  We shall first describe the physical aspects of
the robot in Section~\ref{sec:system}. Then, we detail the
mathematical model used to move and locate the robot in space in
Section~\ref{sec:model}. We show the web-based control interface in
Section~\ref{sec:interface}, and introduce the Hop toolkit used to
build the interface in Section~\ref{sec:hop}. Last, we conclude in
Section~\ref{sec:conclusion} and we discuss future works in
Section~\ref{sec:future}.

\section{Our Pedagogical System}
\label{sec:system}

\begin{figure*}[!htb]
\centerline{
\includegraphics[width=120mm]{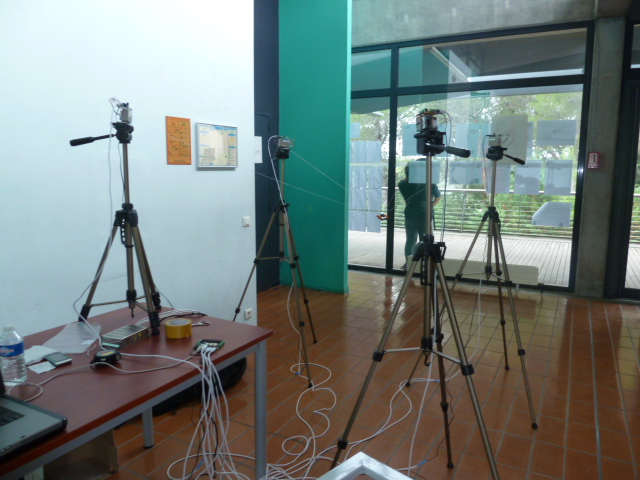}
}
\caption{The cable-driven robot deployed in a classroom}
\label{robot}
\end{figure*}

Our system is pictured in Figure~\ref{robot}. It is composed of four
coiling systems, each put atop of a tripod. The four cables are tied
to a single platform made from Lego. Each coil is activated by
a stepper motor, which is controlled by a Phidget Stepper board. Each
board control interface is plugged by a USB cable. The four USB cables
are plugged on a computer either directly (if there are enough USB
ports), or through a USB hub. It can be any kind of computer with the
USB ``Host'' capability. We made successful experiments with small
embedded systems such as Phidget SBC2 or Fit-PC2 devices. In the
pictured setup, we merely used a standard laptop.

The software controlling the robot was running on the laptop. The
control interface shows two different controls: a manual activation of
each coil to wind or unwind a specific cable, and a Cartesian command
to move the robot in a well-known direction.

By creating a WiFi network in the classroom, or using an existing one,
any connected device could access the Web server running on the laptop
to display the control interface. We actually remotely controlled the
robot from a Nexus S Android device. The WiFi network was also created
by the same Android device, using the connection sharing ability.

This kind of manipulator is portable, modular in size (it can be put
on a table or may cover a whole classroom) and easy to handle by
a teacher or students. Moreover the technologies that are used are
cheap, easily available and so simple that students of any level may
build from scratch their own device.

During the science fair, we used the robot to show how Pythagoras'
theorem is needed to make basic movements. The challenge proposed to
the students was to move the robot from a point $P_1$ to a point $P_2$
in the robot's space using only the manual, per-coil commands. Point
$P_1$ was the initial position of the robot's platform, and point
$P_2$ was figured by the hand of a volunteer student.
Another volunteer student used the control interface to move the
robot's platform. Less than half of the students managed to make the
full movement in less than a minute. Some of them resigned.

We then asked the same students to redo the same $P_1$-to-$P_2$
movement using the Cartesian command. They all succeeded in less than
15 seconds. The challenge opened the students' curiosity for the
robot model.

\section{Cable-driven modelling}
\label{sec:model}

\subsection{Cable-driven robot architecture}
\begin{figure}[!htb]
\centerline{
\includegraphics[width=70mm]{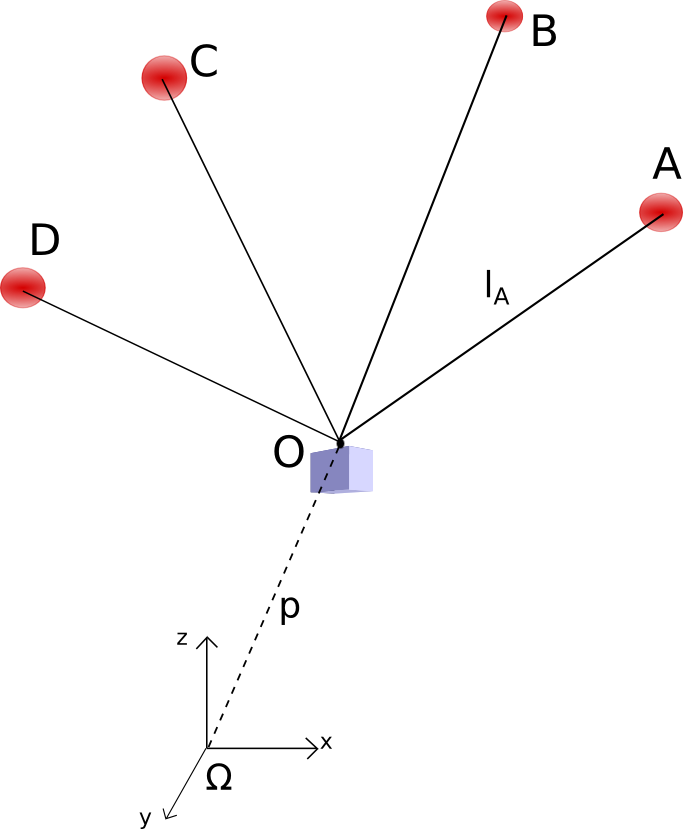}
}
\caption{The cable-driven robot architecture}
\label{robot_arch}
\end{figure}

In the sketch presented in Figure~\ref{robot_arch}, the mobile platform (linked
to the point $O$) is connected to the base (linked to the frame
$\Omega$) by $m=4$ cables.

The cable connects the points $A,B,C$ and $D$ in the base (coordinate $a,b,c$ and $d$ in $\Omega$) to the point $O$ on the mobile platform. The position $p$ of $O$ expressed in $\Omega$ is directly controlled by the length $l_{a,b,c,d}$ and the tension of each cable.

\subsection{Kinematics}
\label{kin}
The implicit kinematics links the position of the platform and the
cable length with the four relations:

\begin{eqnarray*}
||p-a||_2 - l_a &=& 0\\
||p-b||_2 - l_b &=& 0\\
||p-c||_2 - l_c &=& 0\\
||p-d||_2 - l_d &=& 0
\end{eqnarray*}

With these relations, we can compute the four cable lengths to reach
a position for the platform $p$, with the Pythagorean equation:

\begin{eqnarray*}
l_a &=& \sqrt{(x-a^x)^2 + (y-a^y)^2 + (z-a^z)^2}\\
l_b &=& \sqrt{(x-b^x)^2 + (y-b^y)^2 + (z-b^z)^2}\\
l_c &=& \sqrt{(x-c^x)^2 + (y-c^y)^2 + (z-c^z)^2}\\
l_d &=& \sqrt{(x-d^x)^2 + (y-d^y)^2 + (z-d^z)^2}\\
\end{eqnarray*}

\subsection{Cartesian control}
The mobile platform is controlled in the 3 degrees of freedom, its
position, in the global frame $\Omega$.  The inverse kinematics (see
Section~\ref{kin}) allow one to compute the length $l_a$ to reach
a given position.  We compute the order for the motor $A$, denoted
$\rho_a$, as follows:

\begin{equation}
\rho_a = \frac{(l_a - l_0)}{(2\pi r)} N_{steps}
\end{equation}

where $l_0$ is the unwound length of cable at motor's home position, $r$ the drum radius, and $N_{steps}$ the number of steps per turn.
The orders for the other motors $B,C,D$ are computed by the same manner.

\section{Control Interface}
\label{sec:interface}

The interface is generated and served by Hop. We wrote a Web interface
that allows the demonstrator not only to move the robot, but also to
calibrate the coils and get feedback about the robot state. We divided
the interface in three pages: one for the actual control of the robot,
and two for the calibration part: the trilateration of coil
coordinates and zeroing of each motor.

On top of each page of the Web app, links enable to switch between the
different roles or to switch between French and English interface
languages. In this section we present the interfaces, starting with
the Setup and Calibration interfaces. Each interface is composed of
several blocks of information. Each block has a title and can be
folded or unfolded. To reduce the size of the screenshots presented in
this section, some blocks are presented in their folded form.

\subsection{Robot Setup}

\begin{figure}[!htb]
\centerline{\includegraphics[scale=.7]{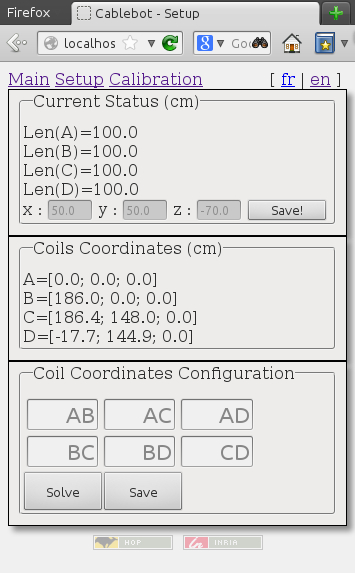}}
\caption{The configuration interface}
\label{applisetup}
\end{figure}

The setup interface is pictured in Figure~\ref{applisetup}. The page
is divided in three blocks. The first block, entitled ``Current
Status'', is shared among all interfaces. It reports the current
cable lengths and the position deduced from the lengths by solving the
direct kinematics. The ``Save!'' button records the
current position in order to get back there later.

The second block reports the last known coils coordinates. It helps to
ensure that the known coordinates roughly correspond to the actual
spatial configuration of the robot. The coordinates are stored within
the application configuration for easy reuse in
subsequent demonstrations.

\begin{figure}[!htb]
\centerline{\includegraphics[scale=.7]{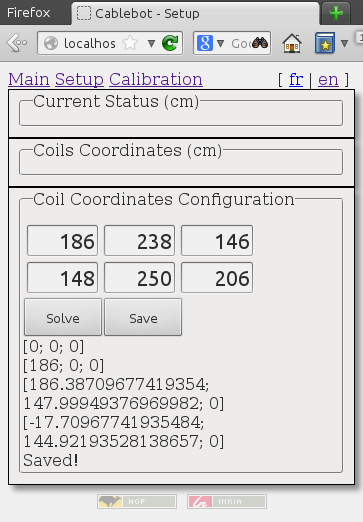}}
\caption{Computing coils coordinates}
\label{applitrilat}
\end{figure}

The third and last block computes the coordinates of the coils using
the coil inter-distances. The Cartesian coordinates are determined by
solving circle equations. The computation takes six inputs (as we have
four coils); two buttons enable solving and saving the trilateration.
Figure~\ref{applitrilat} shows the solution coordinates of a given set
of distances.

\subsection{Robot Calibration}

\begin{figure}[!htb]
\centerline{\includegraphics[scale=.7]{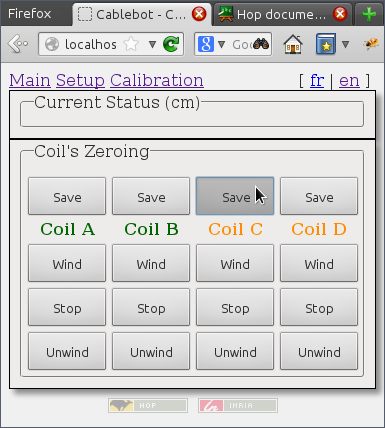}}
\caption{The calibration interface}
\label{applicalib}
\end{figure}

Figure \ref{applicalib} shows the calibration interface, which zeroes
the motors. For each coil, two buttons permit to wind and unwind the
cable until it reach the expected length for the coil (each cable has
a mark at exactly $l_0=100$~cm of the platform). When winding or
unwinding, the coil rotates continuously until the ``Stop'' button is
pressed. The ``Save'' button records the zero (home position) within
the stepper controller. Each coil can be zeroed independently from the
others.  The color tells the user about one coil's motor status: green
stands for zeroed, orange for not zeroed yet, and red signals an
error.  Errors can mean that the motor cannot be detected, or the
system failed to communicate with the motor controller.

\subsection{Robot Control}

\begin{figure}[!htb]
  \centerline{\includegraphics[scale=.7]{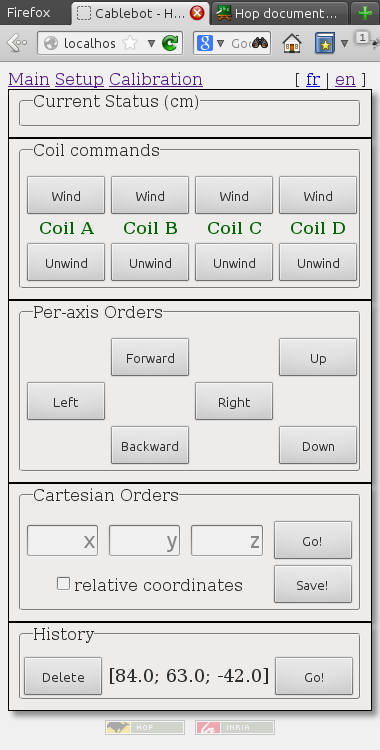}}
  \caption{The main interface}
  \label{applimain}
\end{figure}

The main interface is pictured in Figure \ref{applimain}. It is
composed of five blocks. As the first block is common to previous
interfaces, we omit it here. The second block controls coils. The
colors have the same purpose described above. For each coil there are
two buttons: ``Wind'', and ``Unwind''. Each respectively winds and unwinds
a coil by a half-turn which roughly corresponds to 3.5~cm.

The third block enables per-axis motion: each button moves the robot
in one direction by 5~cm. The fourth block enables the manipulator to
specify coordinates to go to. Either the coordinates represent
a vector by which the robot should be shifted, or, if the checkbox is
unmarked, an absolute position within the coil-delimited affine space.

Last, the fifth control block lists saved coordinates from the status
block. For each saved position, a button loads the coordinates and
moves the robot there, while the second button removes the coordinates
from the list.

\section{Hop}
\label{sec:hop}

Hop is a programming language and platform for the
Web. It incorporates all the required Web-related features into
a single language with a single homogeneous development and execution
platform, thus uniformly covering all the aspects of a Web
application: client-side, server-side, communication, and access to
third-party resources. Hop embodies and generalizes both HTML and
JavaScript Functionalities in a \textsc{scheme/clos}-based platform
\cite{scheme:r5rs} that also provides the user with
a fully general algorithmic language. Web services and APIs can be
used as easily as standard library functions on the server and client.
In this section, we give a brief introduction to Hop. Readers
interested in extra details should refer to \cite{serrano:coordination09, sgl:dls06}.

We use three different aspects of Hop in our system. First, the
ability to interact with system (C/C++) libraries, like the libphidget
one. Second, the Web service definitions: how can we expose a library
function through a Web service, how to deal with concurrency. Last,
the Hop ability to generate interactive Web interfaces, which can
invoke the just described Web services.

\subsection{Layered Control}

The coils are controlled by Phidget boards. The Hop's Phidget library
allows the management of Phidget devices as any other Hop value (such as {\tt String} or any number type). As
such they can be passed to functions, returned as results, or stored
in variables and data structures.

Thanks to the Hop's Phidget library, we wrote all the control code in
Hop. The code is like C-code, with mainly syntactic differences, and
a few stylistic changes. The control code orders a given stepper motor
to rotate in a given direction for a given number of steps.

Atop of this low-level control layer, comes the geometric layer,
translating points in space into cable lengths ($l_{a,b,c,d}$), and
desired cable lengths on a given coil into rotation orders of a given
number of steps ($\rho_{a,b,c,d}$). We choose this separation to keep
the ability to replace the control layer by a simulation layer to draw
or report what would be done for a given position. It enables us to test the geometric layer separately from the control layer.
This separation also allows us to change the geometric layer to deal with different robot designs.

Then, we use Hop to provide access to any functions of the geometric layer through URLs.
This last layer also manages the possible concurrent accesses.


\subsection{Web services}

In Hop, a service is a regular function which can be called from an
URL. The Hop web server waits for HTTP requests on a given port (8080
by default). When a request is received, the requested resource
identifies a service, and the function associated with the service is
applied to the request parameters.

As any Web server, Hop is by default concurrent. Threads, organized in a pool, handle
requests concurrently for better performance and multitasking. A Hop
server is not necessarily serving only one service at a time! Hop uses
native threads, usually POSIX threads, with preemptive semantics. As
such, services with side-effects might require the use of locks to
avoid races.

This concern applies to all services wrapping functions to move the
robot. Moreover, in our specific situation, the robot should not receive
two orders at the same time, even if the orders come from different
services. Here we need a global lock for any movement-related service.
Querying the robot state is side-effect free and thus does not
require resource acquisition.

\subsection{Web interfaces}

Nowadays, Web interfaces are the most universal way to deploy
graphical user interfaces (GUIs). Smartphones and tablets now embed
the same modern browsers as we have on our computers. Web applications
do not require any installation: the application is downloaded from
a Web server and is immediately ready to run. We found it obvious that
a pedagogical system should be as device-agnostic as possible, and not
requiring any complex setup.

Hop is designed to create Web interfaces. Hop embeds Web-UI languages:
it can manipulate any HTML element as it manipulates any Hop value,
\textit{i.e.}, aggregating the element in data structures, converting
it to String and vice-versa.

As Hop defines HTML trees and Web
services, Hop naturally provides simple syntax to call a Hop's service
from a HTML tree, using standard Web events such as mouse clicks or
keyboard inputs.

\section{Conclusion}
\label{sec:conclusion}

We described a pedagogical system made of a cable-driven robot and
a remote control. The robot only requires casual hardware for less
than 600 Euros.  Our system enabled us to give a physical illustration
of the Pythagorean theorem and led to a more attractive presentation
of lectures.  Mobility through wireless remote control improves the
teaching conditions by enabling all students to move the robot.
These positive results motivate us to distribute our robot in
3 different schools (high school, university and engineering school)
during the next year.

\section{Future Works}
\label{sec:future}

There is an ongoing work on Hop called HipHop \cite{bns:plastic11}.
HipHop is about porting the Esterel Reactive Programming style in Hop.
Robots are reactive systems that react to inputs coming from their
environment or from the manipulator.  HipHop is an orchestration
language for Hop. By adding such a layer in the cable-driven robot, we
will be able to easily program robots movement sequences, preempt
actions when an obstacle is detected, and so on.  We plan to place
a smartphone in the robot to get data from its sensors through Hop
(Hop has already been successfully ported to Android platforms).
Using HipHop to orchestrate robot movements, we extend the pedagogical
system to teach robots and event-based programming.  In the robotics
field, we plan to apply our recent work on cable-driven robot
calibration \cite{alexandre_calibration_2012} to this prototype and
add it in our Hop based control interface.

\small
\bibliographystyle{abbrv}
\bibliography{CAR2013}

\end{document}